\documentclass{article}

\usepackage{times}
\usepackage{latexsym}
\usepackage{amsmath}
\usepackage{multirow}
\usepackage{url}
\usepackage{color}
\usepackage{graphicx}
\usepackage{amsmath}
\usepackage{amsfonts}
\usepackage{amssymb}
\usepackage{algorithm}
\usepackage{algpseudocode}
\usepackage{comment}
\usepackage{verbatim}
\usepackage{bm}

\DeclareMathOperator*{\argmax}{arg\,max}

\newcommand{\z}{\mathbf{z}}
\newcommand{\Z}{\mathcal{Z}}
\newcommand{\w}{\mathbf{w}}

\algblock{ParFor}{EndParFor}
\algnewcommand\algorithmicparfor{\textbf{parfor}}
\algnewcommand\algorithmicpardo{\textbf{do}}
\algnewcommand\algorithmicendparfor{\textbf{end\ parfor}}
\algrenewtext{ParFor}[1]{\algorithmicparfor\ #1\ \algorithmicpardo}
\algrenewtext{EndParFor}{\algorithmicendparfor}

\begin{document}

\title{Scalable Probabilistic Entity-Topic Modeling}

\author{
Neil Houlsby\footnote{Work carried out during an internship at Google.}\\
{Department of Engineering}\\
{University of Cambridge, UK}\\
\texttt{nmth2@cam.ac.uk}
\and
Massimiliano Ciaramita\\
{Google Research}\\
{Z\"{u}rich, Switzerland}\\
\texttt{massi@google.com}
}

\maketitle

\begin{abstract}
We present an LDA approach
to entity disambiguation. Each topic is
associated with a Wikipedia article and topics generate either
content words or entity mentions.
Training such models is challenging
because of the topic and vocabulary size, both in the
millions. We tackle these problems using
a novel distributed inference and representation
framework based on a parallel Gibbs sampler guided by the Wikipedia
link graph, and pipelines of MapReduce allowing
fast and memory-frugal processing of large datasets.
We report state-of-the-art performance on a public dataset.
\end{abstract}

\section{Introduction}
Popular data-driven unsupervised learning techniques such as
topic modeling can reveal useful structures in
document collections. However, they
yield no inherent \emph{interpretability} in the structures revealed.
The interpretation is often left to a post-hoc inspection of the
output or parameters of the learned model.
In recent years an increasing amount of work has focused on the task
of annotating phrases, also known as \emph{mentions},
with unambiguous identifiers, referring to \emph{topics},
\emph{concepts} or \emph{entities}, drawn from large repositories such
as Wikipedia.
Mapping text to unambiguous references provides a first
scalable handle on long-standing problems such as
language \emph{polysemy} and \emph{synonymy}, and more generally
on the task of \emph{semantic grounding} for language
understanding.
Resources 
such as Wikipedia, 
Freebase 
and YAGO 
provide enough coverage to support the investigation of Web-scale
applications such search results clustering~\cite{Scaiella:2012}. 

By using such a notion of
\emph{topic} one gains the advantage over pure
data-driven clustering, in that the topics have an identifiable
transparent semantics, be it a person or
location, an event such as earthquakes or the ``financial crisis of
2007-2008'', or more 
abstract concepts such as friendship, expressionism etc. Hence, one
not only gets human-interpretable insights into
the documents directly from the model, but also from a
`grounded interpretation' which allows the
system's output to be interfaced with downstream systems or
structured knowledge bases for further inference.
The discovery of such topics in documents is known
as entity annotation.

The task of annotating entities in documents typically involves two
phases. First, in a \emph{segmentation step}, entity mentions are
identified.  
Secondly, in the \emph{disambiguation} or \emph{linking} step, the
mention phrases are assigned one Wikipedia identifier (alternatively from
Freebase, YAGO etc.).
In this paper we focus upon the latter
task which is challenging due to the enormous space of possible
entities that mentions could refer to. Thus, we assume that the
segmentation step has already been performed, for example
 via pre-processing
the text with a named entity tagger. We then take a probabilistic
topic modeling approach to the mention disambiguation/linking task.

Probabilistic topic models, such as Latent Dirichlet
Allocation (LDA)~\cite{blei2003}, although they do not normally address the
interpretability issue, 
provide a principled, flexible and extensible framework for
modeling latent structure in high-dimensional data.
We propose an approach, based upon LDA,
to model Wikipedia topics in documents. Each topic is
associated with a Wikipedia article
 and can generate either
content words or explicit entity mentions.
Inference in such a model is challenging
because of the topic and vocabulary size, both in the
millions; furthermore, training and applications require the ability to
process very large datasets.

To perform inference at this scale we propose
a solution based on stochastic variational inference (SVI) and
distributed learning. We build upon
the hybrid inference scheme of~\cite{mimno2012} 
which combines document-level Gibbs
sampling with variational Bayesian
 learning of the global topics, resulting in
an online algorithm that yields parameter-sparsity and a manageable
resource overhead. 
We propose a new learning framework that combines
online inference with parallelization, whilst
avoiding the complexity of asynchronous training architectures.
The framework uses a novel, conceptually simple, MapReduce pipeline for
learning; all data necessary for inference (documents, model,
metadata) is serialized via join operations so that each document
defines a self-contained packet for inference purposes. 
Additionally, to better identify document-level consistent topic
assignments, local inference is guided by
the Wikipedia link graph.

The original contributions of this work include:
\begin{enumerate}
\item A large scale topic
 modeling approach to the entity disambiguation task
 that can handle millions of topics as necessary.
\item A hybrid inference scheme that exploits the advantages
of both stochastic inference and distributed processing to achieve
computational and statistical efficiency.
\item A fast Gibbs sampler that exploits model sparsity and incorporates knowledge from the Wikipedia graph directly.
\item A novel, simple, processing pipeline that yields
  resource efficiency and fast processing that can be applied to other
  problems involving very large models.
\item We report state-of-the-art
  results in terms of scalability of LDA models and in
  disambiguation accuracy on the Aida-CoNLL dataset~\cite{Hoffart:2011}.
\end{enumerate}

The paper is organized as follows:
Background on the problem and related work is discussed in the
following section. 
Section~\ref{sect:lda} introduces our model.
Section~\ref{sect:learning} describes the inference scheme, and
Section~\ref{sect:flume} the distributed framework.
Experimental setup and findings, are presented in
Section~\ref{sect:experiments}. 
Follow conclusions.

\section{Related Work}\label{sect:talk}
\begin{figure} \centering
\includegraphics[scale=0.5]{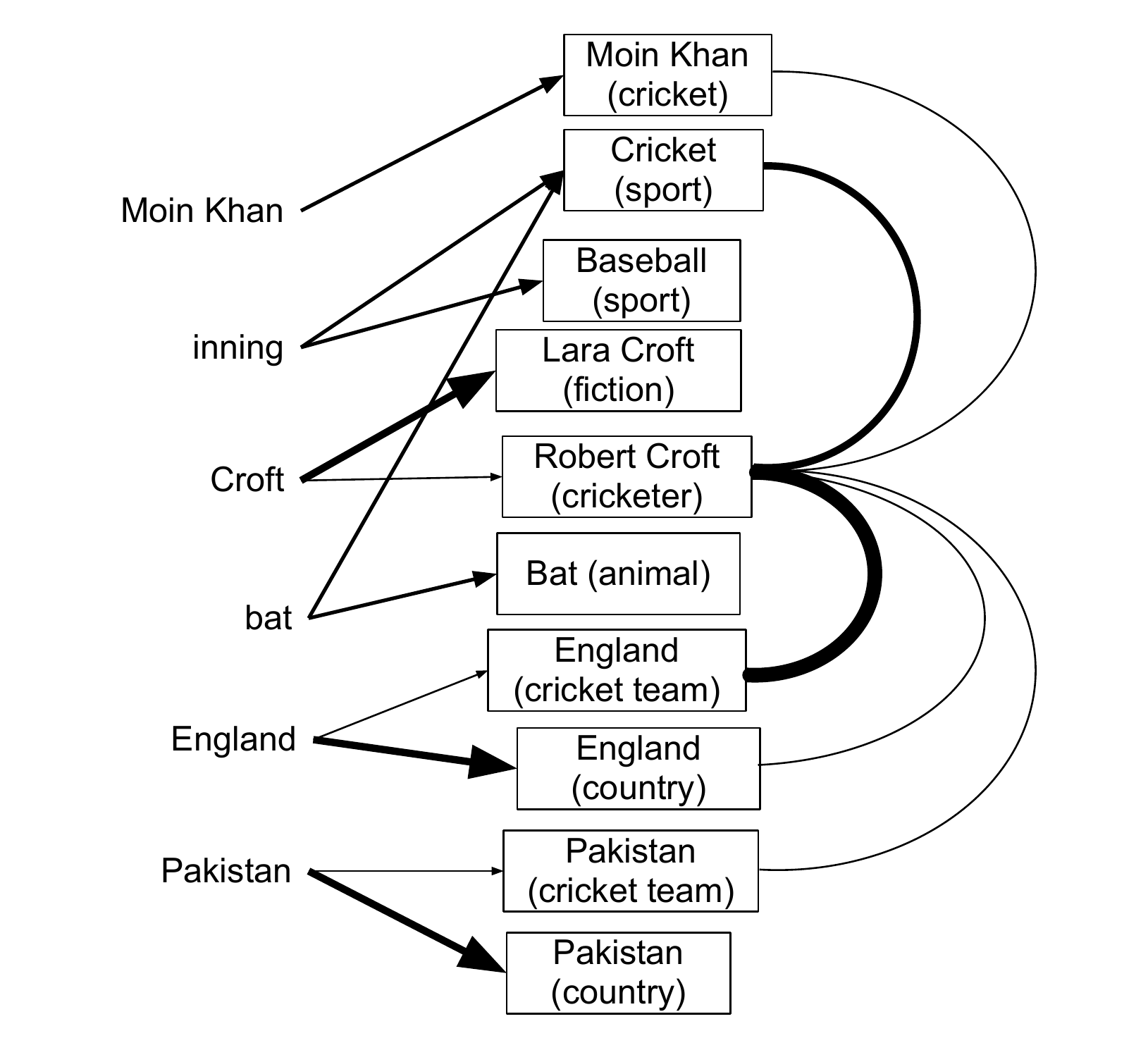}
\vskip-0.5cm
\caption{Example  of document-Wikipedia graph.}
\label{fig:croft}
\end{figure}
Much recent work has focused on
associating textual mentions with Wikipedia
topics~\cite{Mihalcea:2007,Milne:2008,Kulkarni:2009,Ferragina:2010,Hoffart:2011,ratinov2011local,
han-sun:2012:EMNLP-CoNLL}. The
task is known as \emph{topic annotation}, \emph{entity linking}  
or \emph{entity disambiguation}. 
Most of the proposed solutions exploit two
sources of information compiled from Wikipedia: the
link graph, used to infer similarity measures
between topics, and anchor text, to estimate how likely a
string is to refer to a given topic.

Figure~\ref{fig:croft} illustrates the main intuitions behind most annotators'
designs. The figure depicts a few words and names
from a news article about cricket. Connections between strings and
Wikipedia topics are represented by arrows whose line weight
represents the likelihood of that string being used to mention the
topic. In this example, a priori, it is more likely that
``Croft'' refers to the fictional character rather than the
cricket player. However, a similarity graph induced from
Wikipedia\footnote{The similarity measure is typically 
  symmetric.} would reveal
that the cricket player topic is actually densely connected
to several of the candidate topics on the page, those related to
cricket (line weight represents again the connection strength). 
Virtually all topic annotators propose different ways of
exploiting these
ingredients.

A few topic model-inspired approaches have been
proposed for modeling entities
~\cite{Newman:2006,Kataria:2011,han-sun:2012:EMNLP-CoNLL}.
Early work~\cite{Newman:2006} presents extensions to LDA to model both words
 and entities; however, they treat entities as \emph{strings}, not
 linked to a knowledge base.
\cite{Kataria:2011,han-sun:2012:EMNLP-CoNLL} 
model a document as a collection of topic 
mentions, materializing as words or phrases, with topics being
identified with Wikipedia articles.
Kataria \emph{et al.} in particular investigate the use 
of the Wikipedia category graph as the topology of a hierarchical
topic model.
The main drawback of this proposal is its scalability both
in terms of efficiency and topic coverage;
they prune Wikipedia to a subset of approximately 60k
entities, reporting training times of 2.6 days. Han \& Sun carried out
the largest experiment of this kind, training on 3M Wikipedia
documents (and no graph), reporting training times of one
week with a memory footprint of 20GB on one machine.
Our goal is to provide full Wikipedia coverage and high
annotation accuracy with
reasonable training/processing efficiency.

Scalable inference for topic models is the focus of much recent work. 
Broadly, the main approaches divide into two classes: those that
parallelize inference e.g., via distributed sampling
methods~\cite{wang2009,smola2010architecture}, and stochastic
optimization methods~\cite{hoffman2010}.
Computing infrastructures like
MapReduce~\cite{Dean:2008} allow processing of huge amounts of data
across thousands of machines. Unlike in previous work,
we deal with an enormous topic space  
as well as large  datasets and vocabularies. Very high
dimensional models, that can also grow as new data is presented,
impose additional 
constraints such as a large memory footprint, limiting the resources
available for distributed processing.
One solution is to store a model in scalable distributed storage
systems such as Bigtables~\cite{Chang:2008}, and allow individual
processes to read the parameters needed to
process subsets of the data from the global model. This approach
allows individual workers to send back model updates thus
supporting asynchronous training. The downside is the
cost of the worker-model communication which can become
prohibitive and difficult to optimize. Sophisticated asynchronous
training strategies and/or dedicated control architectures are necessary to
address these
issues~\cite{smola2010architecture,Hall:2010:nips,mcdonald-hall-mann:2010:NAACLHLT,Le:2012}.
Recent work on SVI provides an online alternative
to parallelization~\cite{hoffman2010}, this approach can yield memory
efficient inference and good empirical convergence.
In this paper we combine a sparse SVI approach with a distributed
processing framework that gives us massive scalability with our models.


\section{Wikipedia-Topic Modeling}\label{sect:lda}
\subsection{Problem statement}
We follow the task formulation, and evaluation framework, 
of \cite{Hoffart:2011}. 
Given an input text where entity mentions have been identified by a
pre-processor, e.g., a named entity tagger, the goal of a system is to
disambiguate (link) the entity mentions with respect to a Wikipedia
page. Thus, given a snippet of text such as ``[Moin Khan] returns to
lead [Pakistan]'' where the NER tagger has identified the entity mentions
``Moin Kahn'' and ``Pakistan'', the goal is to assign the cricketer id
to the former, and the national cricket team id to the
latter.\footnote{Respectively, \url{en.wikipedia.org/wiki/Moin_Khan}
  and
  \url{http://en.wikipedia.org/wiki/Pakistan_national_cricket_team}.}
We refer to the words outside entity mentions, e.g., ``returns'' and
``lead'', as \emph{content words}.

\subsection{Notation}
Throughout the paper we use the following notation conventions.
\subsubsection{Data}
The training data consists of a collection of $D$ documents,
$\mathcal{D}=\{\w_d\}_{d=1}^D$. 
Note that this data can be \emph{any} corpus of documents, e.g., news,
web pages or Wikipedia itself. 
Each document is represented by a set of
$L_d^c$ content words $\w_d^c = \{w_1^c,\ldots,w^c_{L_d^c}\}$ and
$L_d^m$ entity mentions $\w_d^m = \{w^m_1,\ldots,w^m_{L_d^m}\}$. 

Each word is either a
`content word' or an `entity mention', these two types are distinguished with
a superscript $w^{c},w^{m}$ respectively;  when unambiguous,
this superscript is dropped for readability. Content words consist
of all words occurring in the English Wikipedia articles. Mentions
are phrases (i.e., possibly consisting of several words) that can be
used to refer directly to particular entities e.g. ``JFK Airport'',
``Boeing 747''. Mention phrases are collected from Wikipedia titles,
redirect pages and anchor text of links to other Wikipedia pages.
The vocabularies of words and mentions have size $V^c,V^m$
respectively.
More details on the data pre-processing step are provided in
Section~\ref{sect:data}.

\subsubsection{Parameters}
Associated with each document are two sets of latent variables, 
referred to as `local' parameters because they model only the
particular document in question. The first is the topic assignments for
each content word in the document $\z_d^c=\{z_1,\ldots,z_{L_d^c}\}$,
and the topic assignments for
each entity mention $\z_d^m=\{z_1,\ldots,z_{L_d^m}\}$.
Thus, content words and entity mentions are generated from the same topic
space, each $z_i$
indicates which topic the word $w_i$ (content or mention) is assigned to, where
the topic represents a single Wikipedia entity. For example,
if $w^m_i=\text{``Bush''}$, then $z_i$ could label this word with the topic
``George Bush Sn.'', or ``George Bush Jn.'', or ``bush (the shrub)'' etc.
The model must decide on the assignment based upon the context in which $w^m_i$
is observed. 

The second type of local parameter is the document-topic distribution
$\theta_d$. There is one such distribution per document, and it
represents the topic mixture over the $K$ possible topics that
characterize the document. Formally, $\theta_d$ 
is a parameter vector for a $K$-dimensional multinomial distribution over
the topics in document $d$. For example, in an article about The Ashes\footnote{\url{http://en.wikipedia.org/wiki/The_Ashes}.}, $\theta_d$
would put large mass upon topics such as ``Australian cricket team'',
``bat (cricket)'' and ``Lords Cricket Ground''. 
Note that, although mention and content word's topic assignments are
generated independently, conditioned on the the topic mixture
$\theta_d$, they become dependent when we marginalize out
$\theta_d$, as explained in Section \ref{sect:learning} in more detail. 

There are two vectors of `global' parameters per topic, the `topic-word'
and `topic-mention' distributions $\phi_k^c,\phi_k^m$ respectively.
These distributions represent a probabilistic `dictionary'
 of content words/ mentions associated with the Wikipedia entity
represented by the topic. The content and mention distributions are
essentially treated identically, the only difference being that they are
distributions over different dictionaries of words. Therefore, for
clarity we omit the superscript and the
 following discussion applies to both types.
For, each topic $k$, $\phi_k$ is the parameter vector
of a multinomial distribution over the words, and
will put high mass on words associated with the entity represented by topic $k$.
Because each topic
corresponds to a Wikipedia entity, the number of
 topic-word distributions, $K$,
is large ($\approx4\cdot10^6$);
this provides additional computational challenges not normally
encountered by LDA models.

\subsubsection{Variational Parameters}
When training the probabilistic model, we learn the topic distributions $\phi_k$
from the training data $\mathcal{D}$; again, note that
training is \emph{unsupervised} and $\mathcal{D}$ is any collection of
documents. 
Rather than learn a fixed set of topic distributions, we represent
statistical uncertainty by learning a probability distribution over these
global parameters.
Using variational Bayesian inference (detailed in Section \ref{sec:vb})
 we learn a Dirichlet distribution
over each topic distribution, $\phi_k \sim \text{Dir}(\lambda_k)$, and learn
the parameters of the Dirichlet, $\lambda_k\in\mathbb{R}^V$, called the `variational parameters'.
The set of all vectors $\lambda_k$ represents our model.
Intuitively, each element $\lambda_{kv}$ governs the prevalence
of vocabulary word $v$ in topic $k$; for example, for the topic ``Apple Inc.''
$\lambda_{kv}$ will be large for words such as ``phone'' and ``tablet''.
Most topics will only have a small subset of words from
the large vocabulary associated with them i.e. the topic distributions
are \emph{sparse}. However, the model
would not be robust if we ruled out all possibility of
a new word being associated with a particular topic - this
would correspond to having $\lambda_{kv}=0$. Therefore,
each variational parameter takes at least a small minimum value $\beta$
(defined by a prior, details to follow). Due to the sparsity,
most $\lambda_{kv}$ will take this minimum value $\beta$.
Therefore, in practice, 
to save memory we represent the model using `centered' variational parameters,
$\hat{\lambda}_{kv}=\lambda_{kv}-\beta$, most of which will take value zero,
and need not be explicitly stored.
 
\subsection{Latent Dirichlet Allocation}
\begin{figure} \centering
\includegraphics[scale=0.45]{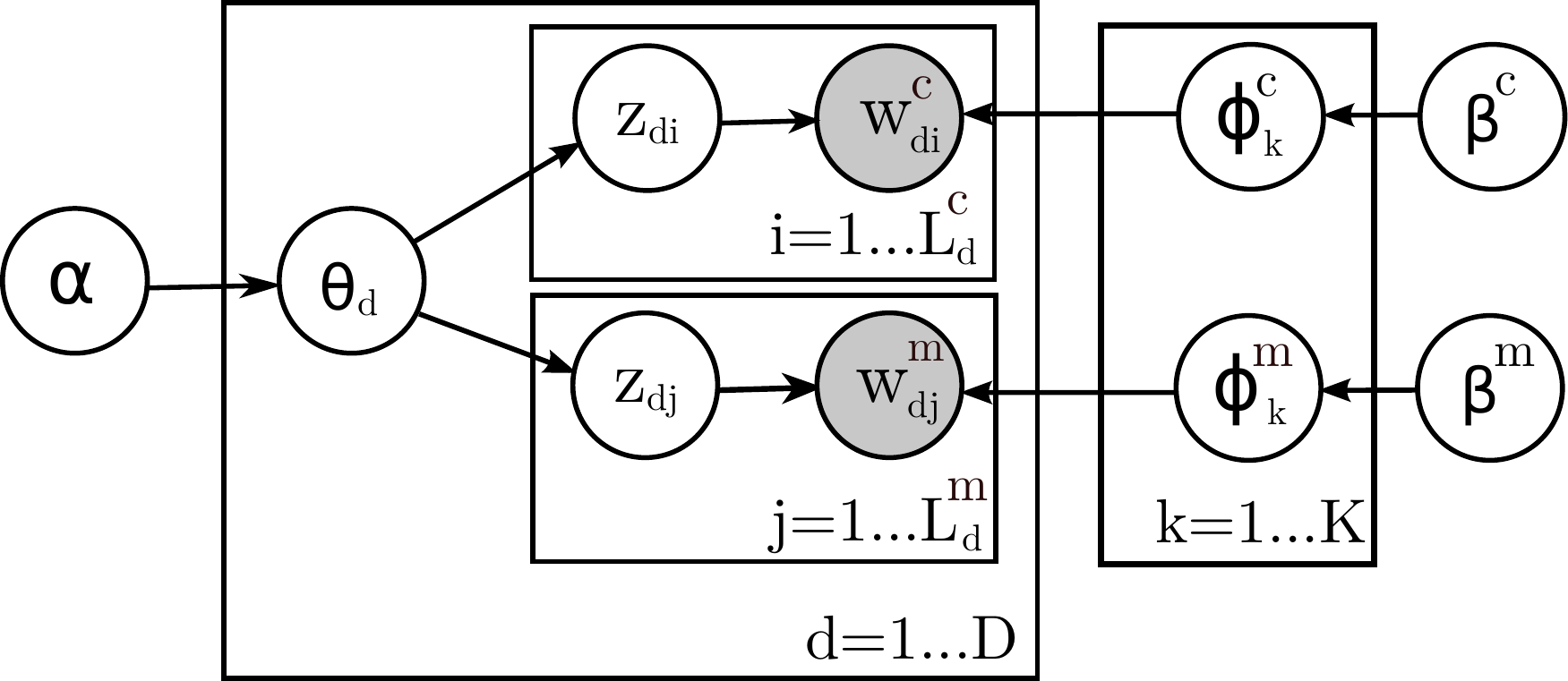}
\caption{LDA with content words (superscript $c$) and mentions (superscript $m$).}
\label{fig:LDAmodelentities}
\end{figure}

The underlying framework for our model is based upon LDA,
a Bayesian generative probabilistic
model, commonly  
used to model text collections~\cite{blei2003}.
We review the generative process of our model below.
The only difference to vanilla LDA is that both mentions and content words
are generated (in the same manner), whereas LDA just considers words.
\begin{enumerate}
\item For each topic $k$ (corresponding to a Wikipedia article), sample
a distribution over the vocabulary of words from a Dirichlet prior
 $\phi_k\sim\text{Dir}(\beta)$.
\item For each document $d$
sample a distribution over the topics
from a Dirichlet prior $\theta_d\sim\text{Dir}(\alpha)$.
\item For each content word $i$ in the document:
\begin{enumerate}
\item Sample a topic assignment from the multinomial: $z_i\sim\text{Multi}(\theta_d)$. 
\item Sample the word from the corresponding topic's word distribution 
$w^c_i\sim\text{Multi}(\phi^c_{z_i})$.
\end{enumerate}
\item For each mention phrase $j$ in the document:
\begin{enumerate}
\item Sample a topic assignment from the multinomial: $z_j\sim\text{Multi}(\theta_d)$. 
\item Sample the word from the corresponding topic's mention distribution 
$w^m_j\sim\text{Multi}(\phi^m_{z_j})$.
\end{enumerate}
\end{enumerate}
Since topics are identified with
Wikipedia articles, we can use the topic assignments to
annotate entity mentions.
$\alpha,\beta$ are scalar hyperparameters for the symmetric Dirichlet priors;
they may be interpreted as topic and word `pseudo-counts' respectively.
By setting them greater than zero, we allow some residual probability
that any word can be assigned to any topic during training.

Documents can be seen as referring to topics either with content
words, e.g., the topic ``Barack Obama'' is 
likely to be relevant in a document mentioning words
like ``election'' ``2012'', ``debate'' and ``U.S.'', but also via
explicit mentions of the entity names such as ``President
Obama'' or ``the 44th President of the United States''. 
It is important to notice that mentions, although to a less degree
than words, can be highly ambiguous; e.g., there
at least seven different ``Michael Jordan''s
in the English Wikipedia -- including two basketball players.

Mentions in text can be detected by running a named entity tagger
on the text, or by heuristic means
~\cite{han-sun:2012:EMNLP-CoNLL}. Here we adopt the former approach
which is consistent with the evaluation data used in our
experiments.\footnote{Off-the-shelf taggers run
typically in linear time with respect to the document length,
thus do not add complexity.}
Thus, a mention is a portion of text
identified as an entity by a named entity tagger.\footnote{We disregard the
label predicted by the tagger.}

However, it is not known to which entity a particular mention
refers and the resolution of this ambiguity is the disambiguation/ linking task.
Assuming the segmentation of the document is known,
the simplest possible extension to LDA to
include topic mentions, derived from Link-LDA~\cite{Erosheva:2004}, is
depicted in Figure~\ref{fig:LDAmodelentities},
and the generative process corresponding to this graph is
outlined above.

Importantly, note that although the topics for each word type are 
sampled independently, their occurrence is coupled
across words and mentions via the document's
topic distribution $\theta_d$. 
During inference, topics appearing in a document
that correspond to content words and those corresponding to mentions
will influence each other. Therefore, during training,
the parameters of the topics $\phi_k^c,\phi_k^m$ 
can learn to capture word-mention co-occurrence.
This enables our model to use the content
words for disambiguating annotations of the mentions.
This sets our approach apart from many current approaches
to entity-disambiguation, which often ignore the content words.

Because the locations of the mentions in a document are
observed, the inference process is virtually identical to LDA.
For ease of exposition, throughout
the paper we present our framework using vanilla LDA,
but extension to the model in Figure \ref{fig:LDAmodelentities}
 is straightforward.
 

\section{Inference and Learning}{\label{sect:learning}}

The model is trained in an unsupervised manner on a corpus
of unlabeled text, e.g. news articles, web-pages, or the Wikipedia
articles themselves. Only during initialization of the model
do we use supervised information from Wikipedia articles, which
by construction of our model, are each labeled with a single topic.

The English Wikipedia contains around 4M articles (topics). The
vocabulary size for content words and mention strings is, respectively,
around 2M and 10M.
Given the vast potential size of the parameter space (topic-word, and
topic-mention matrix), learning a sparse set of parameters is
essential, and large corpora are required, necessitating
a highly scalable framework.

\subsection{Hybrid inference}
We build upon a hybrid variational inference and Gibbs sampling
framework~\cite{mimno2012}.
The key advantages of this method are statistical efficiency
from the online variational inference (the parameters
are updated online,
before waiting for all the data to be processed), and parameter sparsity from taking
finite samples.
Here we present the key equations, together with reformulations that yield
a fast implementation of the sampler.
For notational brevity we present the equations in this section for 
content word
modeling only (i.e. omitting $w^m_{di}, \phi_k^{m}$, $\beta^{m}$
nodes and the superscript $c$. 
from Figure \ref{fig:LDAmodelentities}); given the conditional independence
assumptions the equations are easily
extensible to the model in Figure \ref{fig:LDAmodelentities}.

\subsection{Variational Bayes} \label{sec:vb}
The goal of learning in LDA is to infer the posterior distribution
of topics $\phi_k$. When performing inference on documents
we seek the `local' topic assignments $\z_d$. 
We integrate/collapse out $\theta_d$, which is found to
improve convergence~\cite{teh2007}. Bayes rule is employed to compute the joint
posterior \\
$p(\z_1,\ldots,\z_D, \phi_1,\ldots, \phi_K | \mathcal{D}, \alpha,
\beta)$.
This computation is not tractable, and hence approximate variational 
Bayesian inference
is used. Variational inference
 involves approximating a complex posterior distribution with a
 simpler one, $q(\z_1,\ldots,\z_D, \phi_1,\ldots, \phi_K)$. The latter is fitted
to the true posterior so as to maximize the `Evidence Lower Bound' (ELBO),
a lower bound on the log marginal probability of the data $p(\mathcal{D}|\text{model})$~\cite{bishop2006}.
We use the following approximation to the posterior:
\begin{align}
p(\z_1,\ldots,\z_D, \phi_1,\ldots, \phi_K | \mathcal{D}, \alpha,
\beta)
&\approx 
q(\z_1,\ldots,\z_D)q(\phi_1,\ldots,\phi_K)\notag\\
&=\prod_d q(\z_d) \prod_k q(\phi_k; \lambda_k)\,. \label{eqn:approx_post}
\end{align}
In \eqref{eqn:approx_post} we assume statistical independence in
the $K$ topics and $D$ topic-assignment vectors.
Importantly, however, independence is not assumed between
the elements of each document's assignment vector, $\z_d$.
The correlations between the topics are key to modeling topic
consistency in the document, and the sparse computations
that follow. 

The ELBO is optimized with respect to the 
variational distributions $q(\z_1,\ldots,\z_D)$, $q(\phi_1,\ldots\phi_K)$
 in an alternating
manner; one distribution is held fixed, while the other is optimized.
The variational distribution over topics $q(\phi_k;\lambda_k)$ is
a Dirichlet distribution,
with $V-$ dimensional parameter vector $\lambda_k$, one for each topic.
The elements of the topic's variational parameter vector $\lambda_{kv}$
give the importance of word $v$ in topic $k$. This can be observed
from the mean
of the Dirichlet, which is the multinomial:
$\mathbb{E}_{\phi_k}[q(\phi_k|\lambda_k)]=p(v|k)=\text{Multi}(\lambda_{kv}/\sum_k\lambda_{kv})$.
The variational parameters $\lambda_k$
are optimized during learning, with $\{q(\z_d)\}_{d=1}^D$ fixed.
The optimal variational parameters are given by:
\begin{equation}
\lambda_{kv}=\beta+\sum_{d=1}^D\sum_{i=1}^{L_d}\mathbb{E}_{q(\z_d)}
[\mathbb{I}_{z_{di}=k}\mathbb{I}_{w_{di}=v}]\,.\label{eq:batch}
\end{equation}
For brevity
we shall henceforth refer to these variational parameters
simply as the `parameters' of the model.

The optimal $q(\z_d)$ given $\{q(\phi_k;\lambda_k)\}_{k=1}^K$ is given by
\begin{align*}
q(\z_d)\propto \exp\{\mathbb{E}_{q(\{\z_{1:D}\setminus \z_d)}[\log p(\z_d|\alpha)p(\w_d|\z_d,\beta)])\}\,,
\end{align*}
where $q(\z_{1:D}\setminus \z_d)$ is the variational distribution
over all assignment vectors for all documents excluding $d$.
However, in stead of parameterizing $q(\z)$ and performing 
variational inference we Gibbs sample from $q(\z)$.
This involves sequentially visiting each topic and sampling
that topic conditioned on all other assignments and the other parameters
of the (variational) posterior and word $w_i$:
$z_i\sim p(z_i|\z^{\setminus i}, \lambda_1,\ldots,\lambda_K, w_i)$.

The key advantage of sampling the assignments is that 
one can retain sparsity.
Most improbable word-topic assignments will not be sampled, 
the result being that many of the elements $\lambda_k$ remain constant.

\subsection{Stochastic variational inference}
The key insight behind SVI is
that one can update the model parameters $\lambda_k$ from
just a subset of the data, $\mathcal{B}$~\cite{hoffman2012}.
This enables one to discard the local variables (sampled topic
assignments) after each update is performed. Having performed
inference on only a subset of the data, 
one achieves only a noisy estimate of the full batch update step;
but, provided that the noisy estimates are unbiased (averaging
over the data sub-sampling process), one can guarantee convergence
to an optima of the full batch solution. 
The correct update scheme is achieved
by interpolating the noisy updated parameters from the subset
 with the old ones:
\begin{equation}
\lambda_{kv} = (1-\rho)\lambda_{kv}^{\text{old}} 
+ \rho\frac{|\mathcal{D}|}{|\mathcal{B}|}\lambda_{kv}^{\text{new}} \label{eq:interp}
\end{equation}

The scaling of the update by $|\mathcal{D}|/|\mathcal{B}|$ ensures
that the expected value of the update is equal to the batch
update that uses all of the data, as required.

The stochastic approach has two key advantages. Firstly, one does not have to wait
until the entire dataset is observed before performing even a single update
to the parameters (as 
in the batch approach, Eq.~\eqref{eq:batch}), which yields improved
convergence.
Secondly, by discarding the local samples after each mini-batch $\mathcal{B}$ one
can save vast amounts of memory. The requirement 
to store and communicate all of the local samples can be prohibitive in
approaches based purely upon Gibbs sampling~\cite{wang2009}. 
Section~\ref{sect:flume} details how the hybrid scheme is incorporated
into a distributed framework.

\subsection{Implementation of sparse sampling}

Beyond sparsity, when working with a very
large topic space it is important to perform efficient Gibbs sampling.
For each word in each document (and for each sweep)
 we must sample $z_{i}$ from a $K$ dimensional
multinomial.
Naive sampling would require
 $\mathcal{O}(K)$ operations. However, if one 
judiciously visits the high probability topics first, the number 
of computations can be vastly reduced, and any $\mathcal{O}(K)$ operations
can be pre-computed.
The sampling distribution for $z_{i}$ is given by (for brevity
we omit the parameters):
\begin{align}
q(z_i=k | \z^{\setminus i}, w_{i}=v) &\propto (\alpha + N_{k\cdot}^{\setminus i})
\exp\{\mathbb{E}_q [\log\phi_{kv}]\}\,, \label{eq:samplingdist} \\
\exp\{\mathbb{E}_q [\log\phi_{kv}]\} &= \exp\{\Psi(\beta + \hat{\lambda}_{kv}) -
\Psi(V\beta + \hat{\lambda}_{k\cdot})\}\,. \notag
\end{align}

$\hat{\lambda}_{kv} = \lambda_{kv}-\beta$ denotes `centered' parameters,
these are initialized to zero for most $k,v$.
$N_{kv}^{\setminus i}=\sum_{j\neq i}\mathbb{I}[z_j=k,w_j=v]$ counts
the number of assignments of topic $k$ to word $v$ in the document,
and the subscript dots in 
$N_{k\cdot},\,\hat{\lambda}_{k\cdot}$ are shorthand for the summation over index
$v$, 
e.g. $N_{k\cdot}^{\setminus i}$ counts total occurrences of topic $k$ in the document.
$\Psi()$, denotes the Digamma function. 
To avoid $\mathcal{O}(K)$ operations we decompose the sampling
distribution as follows:
\begin{align}
q(z_i=k | \z^{\setminus i}, w_{i}=v) \propto 
\underbrace{\frac{\alpha \exp\{\Psi(\beta)\}}{\kappa'_{k}}}_{\mu_k^{(d)}}
+ \underbrace{\frac{\alpha
    \kappa_{kv}}{\kappa'_{k}}}_{\mu_k^{(v)}}+
& \underbrace{\frac{N_k^{\setminus i}\exp\{\Psi(\beta)\}}{\kappa'_{k}}}_{\mu_k^{(c)}}
+ \underbrace{\frac{N_k^{\setminus i}\kappa_{kv}}{\kappa'_{k}}}_{\mu_k^{(c,v)}}
\label{eq:rearrangement}
\end{align}

where $\kappa_{kv} = \exp\{\Psi(\beta + \hat{\lambda}_{kv})\} - \exp\{\Psi(\beta)\}$
and $\kappa'_k = \exp\{\Psi(V\beta + \hat{\lambda}_{\cdot k})\}$
are transformed versions of the variational parameters.
$\mu_k^{(d)}$ is dense, but it can be precomputed.
For each word $\mu_k^{(v)}$ has mass only for the topics for
which $\hat{\kappa}_{kv} \neq 0$; for each word in the document this can be
precomputed. $\mu_k^{(c)}$ has mass only for currently observed topics
in the document, i.e. those for which $N_k^{\setminus i} \neq 0$;
this term must be updated every time we sample, but it can be 
done incrementally. $\mu_k^{(c,v)}$ is non-zero only for topics which
have non zero parameters and counts, 
but must be recomputed for every update and new word.
If most of the
topic-mass is in the smaller components (which can be achieved
via appropriate choices of $\alpha,\beta$), visiting
these topics first when performing sampling will
require much fewer than $\mathcal{O}(K)$ operations.
To compute the normalizing constant of \eqref{eq:samplingdist}
the rearrangement \eqref{eq:rearrangement} is exploited with
$\mathcal{O}(K)$ sums in the initialization, followed by sparse online
updates. 
Algorithm~\ref{alg:gibbs} summarizes the processing of a single
document.

Algorithm \ref{alg:gibbs} receives as input the document $\w_d$, initial
topic assignment vector $\z_d^{(0)}$, and transformed parameters $\kappa'_k,\kappa_{kv}$.
Firstly, the components of the sampling distribution in \eqref{eq:rearrangement}
that are independent
of the topic counts $\mu^{(d)},\mu^{(v)}$ and their corresponding normalizing constants
$\mathcal{Z}^{(d)},\mathcal{Z}^{(v)}$ are pre-computed (lines 2-3).
This is the only stage at which the full dense $K$--dimensional vector
$\mu^{(d)}$ needs to be computed.
Note that one only computes $\mu_k^{(v)}$ for the words in the
current document, not for the entire vocabulary.
Next, at the beginning of each Gibbs sweep, $s$, the counts for each
word-topic pair $N_{kv}$ and overall topic-counts $N_{k\cdot}$ are computed from the
initial vector of samples $\z^{(0)}$ (lines 5-6).
During each iteration
of sampling the first operation is to subtract the current topic from
the counts in line 8. Now that the topic count has changed, the 
two count-dependent components of the sampling
distribution are computed
(note that $\mu_k^{(c)}$ can be updated from the previous sample, but
$\mu_k^{(c,v)}$ must be recomputed for the new word). 
The four
components of the sampling distribution and their normalizing constants are summed
in lines 13-14 and a single topic is drawn for the word at location $i$ (line 15).
The topic-word counter for the current sweep is updated in line 16,
and if the topic has changed since the previous sweep the total topic count
is updated accordingly (line 18).

The key to efficient
sampling from the multinomial in line 15 is to visit
$\mu_{k}$ in order
$\{ k\in\mu_k^{(c,v)},\,k\in\mu_k^{(c)},\,k\in\mu_k^{(v)},\,k\in\mu^{(d)}_k \}$.
A random schedule would require on average $K/2$ evaluations of
$\mu_k$. However, if the distribution is skewed, with most
of the mass in the sparse components, then much
fewer evaluations are required if these topics are visited first. 
The degree of skewness in the
distribution is governed by the initialization of the parameters, and
the priors $\alpha, \beta$. Because the latter act as pseudo-counts,
setting them to small values favors sparsity.

After completion of all of the Gibbs sweeps
the topic-word counts from each sweep $N^{(s)}_{kv}$ are averaged (discarding an
initial burn in period of length $B$) to yield updated parameter
values $\hat{\lambda}_{kv}^d$

\begin{algorithm}
\caption{Inner Gibbs Sampling Loop}
\label{alg:gibbs}
\begin{algorithmic}[1]
\State \textbf{input}: $(\w_d, \z^{(0)}_d, \{\kappa_{kv}\},\{\kappa'_{k}\})$
  \State $\mu^{(d)}_k \gets \alpha e^{\Psi(\beta)}/\kappa'_k,\,\Z^{(d)}\gets\sum_k{\mu_k^{(d)}}$ 
  \State $\mu^{(v)}_k \gets \alpha \kappa_{kv}/\kappa'_k,\,\Z^{(v)}\gets\sum_k{\mu_k^{(v)}}\,\,\forall
  v\in\mathcal{D}$
\For{$s \in 1\ldots S$}  \Comment{{\small Perform $S$ Gibbs sweeps.}}
  \State $N_{kv}^{(s)} \gets \sum_{i=1}^{L_d}
  \mathbb{I}_{z_{i}^{(s-1)}=k}\wedge \mathbb{I}_{w_{i}^{(s-1)}=v}$
		\Comment{{\small Initial counts}}
  \State $N_{k\cdot} \gets \sum_{v|N_{kv}>0}N_{kv}^{(s)}$
  \For{$i \in 1\ldots L_d$}   \Comment{{\small Loop over words.}}
    \State $N_{k}^{\setminus i} \gets N_{k\cdot} - \mathbb{I}_{z_{i}=k}$ \Comment{{\small Remove topic $z_i$ from counts.}}
    \State $\mu^{(c)}_k \gets N_k^{\setminus i}
    e^{\Psi(\beta)}/\kappa'_k$
    \State $Z^{(c)}\gets\sum_k{\mu_k^{(c)}}$
    \State $\mu^{(c,v)}_k \gets N_k^{\setminus i} \kappa_{kw_i}/\kappa'_k$
    \State $\Z^{(c,v)}\gets\sum_k{\mu_k^{(c,v)}}$
    \State $\mu_k \gets \mu_k^{(d)} + \mu_k^{(v)} + \mu_k^{(c)} + \mu_k^{(c,v)}$
    \State $\Z \gets \Z^{(d)} + \Z^{(v)} + \Z^{(c)} + \Z^{(c,v)} $
    \State $z_{i}^{(s)}\sim \mathrm{Multi}(\{\mu_k/\mathcal{Z}\}_{k=1}^K)$
			\Comment{{\small Sample topic.}}
    \State $N_{z_i^{(s)}}^{(s) w_i}\gets N_{z_i^{(s)} w_i}^{s} + 1$ \Comment{{\small Update counts.}}
    \If{$z_{i}^{(s)}\neq z_{i}^{(s-1)}$}
      \State update $N_{k\cdot}$
    \EndIf
  \EndFor
\EndFor
\State $\hat{\lambda}^{d}_{kv} \gets \frac{1}{S-B}\sum_{s>B} N_{kv}^{(s)}$ \Comment{{\small Compute updated parameters.}}
\State \textbf{return}: $\hat{\lambda}^{d}_{kv}$
\end{algorithmic}
\end{algorithm}

After completion of Algorithm \ref{alg:gibbs}, 
the parameter updates from the processing of each document 
$\hat{\lambda}^d_{kv}$
 are interpolated with the current
values on the shard using $\eqref{eq:interp}$ to complete the local SVI procedure.
In practice, we found a baseline local minibatch size $|\mathcal{B}|$ of one, and a
small local update weight $\rho_{\text{loc}}$ already worked well.

\subsection{Incorporating the graph}
Most non-probabilistic approaches
to entity disambiguation achieve good performance by using
the Wikipedia in-link graph. 
We exploit the Wikipedia-interpretability of the topics
to readily include the graph into our sampler.
Intuitively, we would like to
weight the probability of a topic, not only by the presence of other
topics in the document, but by a measure of its consistency
with these topics. This is in line
with the Gibbs sampling approach where, by construction, all
topics assignments, except the one being considered, are known.
For this purpose we introduce the
following coherence score:
\begin{equation}
\mathrm{coh}(z_{k}|w_i) = \frac{1}{|\{\z_d\}|-1} \sum_{k'\in \{\z_d\}^{\setminus i}}
\mathrm{sim}(z_k,z_{k'})\,.
\label{eq:coh}
\end{equation}
where $\{\z_d\}$ is the set of topics induced by assignment $\z_d$ for
document $d$, and $\mathrm{sim}(z_k,z_{k'})$ is the `Google
similarity'~\cite{Cilibrasi:2007,Milne:2008b} 
between the corresponding Wikipedia pages.

We include the coherence score by augmenting
$N_k^{\setminus i}$ in Eqn. \ref{eq:rearrangement} as
$N_{k}^{\setminus i} = (N_{k\cdot} -
\mathbb{I}_{z_{i}=k})\mathrm{coh}(z_{k|w_i})$.
Thus, the coherence contributions is appropriately
incorporated into the computation of the normalizing constant. 
Adding coherence will change the
convergence of the model, however,
we perform a relatively small numbers of Gibbs sweeps;
full convergence is not desired anyway
because it would yield an impractical dense solution.
In practice, the addition of coherence to the sampler proves
effective. 

Previous work has extended LDA to learn topic correlations, for
example,
 by using a more sophisticated prior \cite{lafferty2005correlated}.
Learning the correlations in such a manner, using the Wikipedia
graph for guidance, could provide a effective alternative solution.
However, extending the model in this direction and maintaining scalability
is a challenging problem, and an opportunity for future research.
 Alternatively, our simple approach, that incorporates the graph
directly into the sampler, provides an effective solution
that does not increase the complexity of inference.

\section{Distributed Processing}\label{sect:flume}
We use MapReduce~\cite{Dean:2008} for distributed
processing of the input data. A dataset is 
partitioned in \emph{shards} which are processed
independently by multiple workers. Documents, model
parameters and all other data used for inference are stored in 
\emph{SSTables}~\cite{Chang:2008}, immutable key-value maps
where both keys and values are arbitrary strings. Keys are sorted by
lexicographic order, allowing efficient joins over multiple
tables with the same key types. Values hold serialized
structured data encoded as
\emph{protocol
  buffers}\footnote{\url{https://code.google.com/p/protobuf}}. We
denote such tables as {\tt <K,V>}, where {\tt K} is a
key type and {\tt V} a value type, when the value is a
collection of objects we use the notation {\tt<K,<V>>}.

\subsection{Pipelines of MapReduce} \label{sect:pipeline}

Each worker needs the current model
to perform inference. The model is typically large to
start with and can grow larger as new explicit parameters can be
added after each iteration. Storing huge models 
in memory on many machines is impractical. One
solution is to store the model in a distributed data structure, e.g., a
Bigtable. A shortcoming of using a centralized model is the latency
introduced due to concurrent worker-model communication~\cite{smola2010architecture,Hall:2010:nips,mcdonald-hall-mann:2010:NAACLHLT,Le:2012}.

We propose a novel, conceptually simple, and
practical alternative. Documents are stored in {\tt <I,D>} tables
where the key is a document identifier and the value holds the
document content. The
current model is stored in a table {\tt <V,<T>>} keyed by a symbol (a
mention or content word)
 while the value is a collection of the symbol's topic
parameters.
Before inference we process the data and generate tables
re-keyed by symbols, whose values are the document identifiers of the
documents where the symbol occurred. Then a
\emph{join}\footnote{With the term join we always mean an outer join.}
is performed of the new table with the model table, which outputs a
table {\tt <I,<T>>} keyed by 
document id, whose values are all model parameters appearing in the
corresponding document. A document and its parameters can now
be streamed in the inference step, by-passing
the issue of representing the full model anywhere, either in local memory or
in distributed storage.

Additional meta-data, e.g. the
Wikipedia graph, can be passed to the document-level sampler in a
similar fashion by generating a table {\tt <I,<E>>}
keyed by document 
identifiers, whose values are the edges of the Wikipedia graph connecting
 the topics in that document.
Thus, document-model-metadata tuples which define self-contained blocks
with all information needed to perform inference on one document are
streamed together.
After inference, the topic assignments are outputted. While
training, updates are computed, streamed 
out and aggregated over the full dataset by reducers.
Updates are stored in a table {\tt <V,<T>>} keyed by a 
symbol which can finally be joined with the original model table for
interpolation with old values (as defined by the SVI procedure \eqref{eq:interp}),
to generate a new model.

\begin{figure} \centering
\includegraphics[scale=0.46]{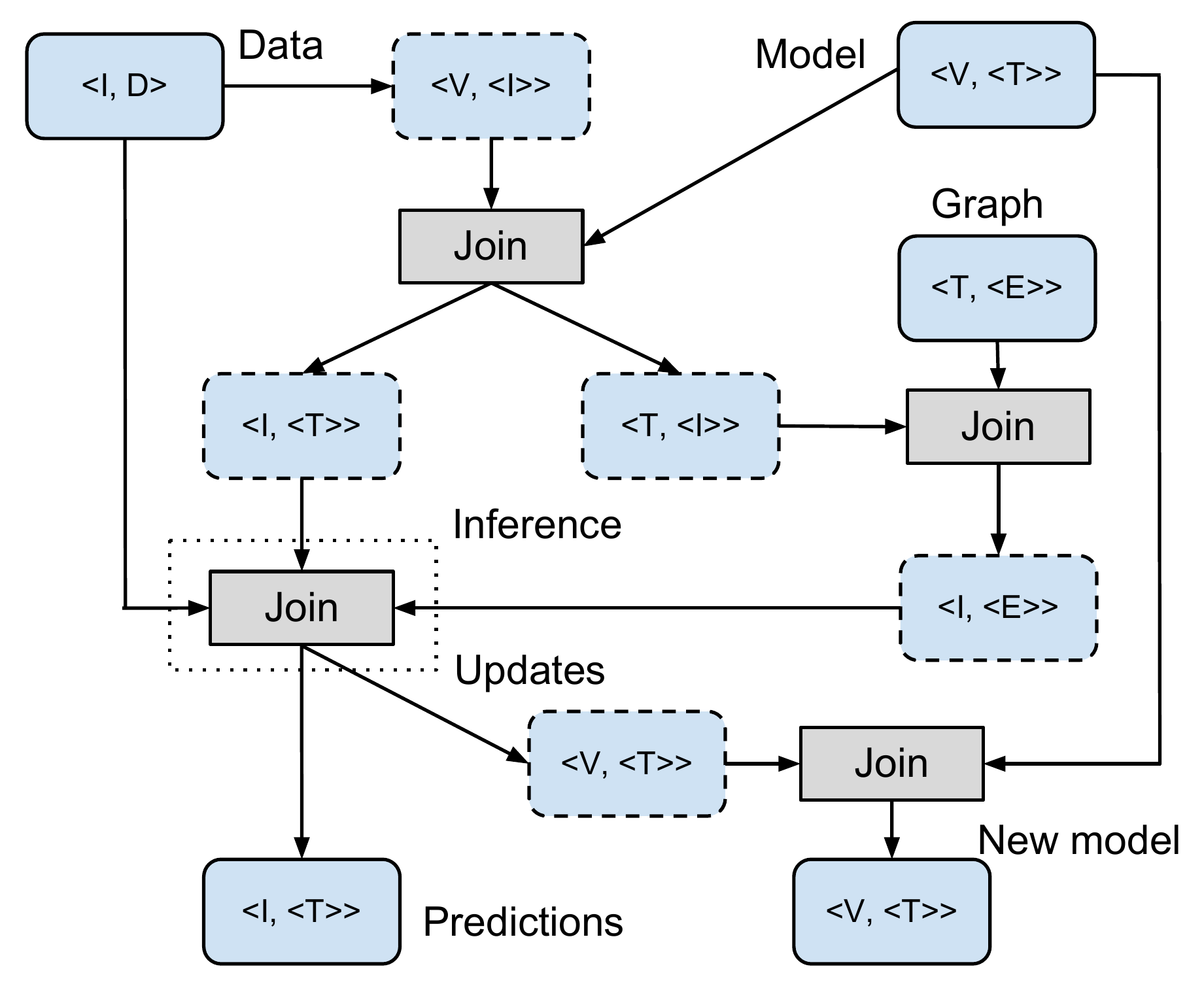}
\caption{Pipeline of MapReduce flow graph.}
\label{fig:flume}
\end{figure}

Figure~\ref{fig:flume} illustrates the flow graph for the
process corresponding to one iteration. Rounded boxes with
continuous lines denote input or output tables,
dashed-line boxes denote intermediate outputs.
Although apparently complex, this procedure is
efficient since the join and data re-keying operations are
faster than the inference step which dominates
computation time. The procedure can
produce large intermediate outputs, however, these are only temporary
and are deleted immediately after being consumed. We implement the
pipeline using Flume~\cite{Chambers:2010} which
greatly simplifies coding and execution, and takes care of the clean-up of
intermediate outputs and optimization of the
flow of operations. The proposed architecture is
significantly simpler to replicate using public
software than complex asynchronous solutions e.g. 
\cite{smola2010architecture,coates2012emergence}.
Open source implementations of Flume and other MapReduce pipelines (e.g.,
Pig~\cite{Olston:2008}) are becoming increasingly
popular and are publicly available, opening up new
opportunities for machine learning at web scale.\footnote{E.g., see
  \url{http://flume.apache.org/.}} 

\subsection{Combined procedure} \label{sect:combined}
The overall procedure for parameter re-estimation takes the
following pathway which is summarized in Algorithm \ref{alg:main}.
Globally we store (on disk, not in memory) just the sparse set of
parameters, and their 
sum over words for each topic (lines 1-2). We perform
$T$ global iterations of Stochastic Variational Inference.
Using the pipeline described in
Section~\ref{sect:pipeline}, parameters (and meta-data)
corresponding to the words in each shard of data (i.e. only those
$\lambda_{kv}$ for which word $v$ appears at least once in $\mathcal{D}^m$) are
copied to an individual worker, along with relevant pre-computed
quantities, $\lambda_{\cdot k}$. 
In lines 6,7 the transformed parameters $\kappa_{kv}, \kappa'_k$
are computed using Eqn.~(\ref{eq:rearrangement}) once at the initialization stage of each worker.
The initial topic assignments are set in line 8.
In each worker we loop sequentially over the documents,
performing Gibbs sampling (algorithm \ref{alg:gibbs}).
We run an `inner loop' of SVI on each shard;
after sampling a single document, the local copy of the model 
parameters $\hat{\lambda}_{kv}$ is updated
using weighted interpolation in line 10. In line 11, the dense
vector $\lambda_{k\cdot}$ is updated incrementally, i.e. its values 
corresponding to topics
that have not been observed in the current document do not change.
After processing all of the documents in the shard, the 
updates from each document are averaged (line 13),
and the global parameter updates
are aggregated and interpolated with the previous model (line 15).
This completes the `outer loop' of SVI. 
The minibatch size for the outer loop SVI
is equal to the number of documents per shard.
We use a minibatch size of one in the inner loop SVI,
as presented in Algorithm~\ref{alg:main}, extension to
arbitrary minibatches is straightforward.

For simplicity we set the interpolation $\rho_{\text{loc}},
\rho_{\text{global}}$ to be constant.
It is straightforward to extend the algorithm to use
a Robbins-Monroe schedule~\cite{robbins1951stochastic}.
Recent work has developed methods for automatic setting of this
parameter \cite{schaul2012no,ranganath2013adaptive}, investigating
an optimal update schedule within our framework is a subject for future work.
\begin{algorithm}
\caption{Distributed inference for LDA}
\label{alg:main}
\begin{algorithmic}[1]
\State {\small initialize $\hat{\lambda}_{kv} \gets$ Wikipedia initialization (Section~\ref{sect:data}).}
\State $\hat{\lambda}_{k\cdot} = \sum_v \hat{\lambda}_{kv}$
{\small \Comment{Dense $K$-dim vector.}}
\For{$t = 1\ldots T$} \Comment{{\small Global SVI iterations.}}
\ParFor{{\small $m=1,\ldots,M; \mathcal{D}^{m}\subset\mathcal{D}$ }}
{\small \Comment{MapReduce.}}
  \For{$d \in \mathcal{D}^m$}
		\State {\small $\kappa_{kv}\gets
  		\exp\{\Psi(\beta + \hat{\lambda}_{kv})\} - \exp\{\Psi(\beta)\}$} \Comment{{\small Sparse.}}
		\State {\small $\kappa'_{k}\gets
 			\exp\{\Psi(V\beta + \hat{\lambda}_{k\cdot})\} $ \Comment{Dense $K-$dim vector.}}
    \State {\small $\z_{d}^{(0)} \gets$ TagMe initialization (Section~\ref{sect:init})}
    \State {\small $\hat{\lambda}_{kv}^d \gets $ Algorithm \ref{alg:gibbs},
						\textbf{input}: ($\w_d ,\z_d^0, \{\kappa_{kv}\},\{\kappa'_k\}$})
		\State {\small $ \hat{\lambda}_{kv} \gets (1 - \rho_{\text{loc}})\hat{\lambda}_{kv} +
			\rho_{\text{loc}}|\mathcal{D}^m|\hat{\lambda}^d_{kv} $}
			\Comment{{\small Update locally.}}
		\State $\hat{\lambda}_{k\cdot} = \sum_v \hat{\lambda}_{kv}$ 
			\Comment{{\small Update incrementally (sparse).}}
  \EndFor
	\State {\small $\hat{\lambda}_{kv}^m \gets \sum_{d\in\mathcal{D}^m}\hat{\lambda}_{kv}^d$}
\EndParFor
\State {\small $\hat{\lambda}_{kv}^{(t)}\gets(1-\rho_{\text{global}})\hat{\lambda}_{kv}^{(t-1)} 
+ \frac{\rho_{\text{global}}}{M}\sum_m\hat{\lambda}_{kv}^m$}
\EndFor
\end{algorithmic}
\end{algorithm}

\section{Experiments}\label{sect:experiments}
For statistical models with a very large parameter space, 
and many local optima, such as the proposed Wikipedia-LDA model,
 the initialization of the parameters
has a significant impact upon performance (deep neural networks are another
classical example where this is the case). Empirically, we found that
in this vast topic space,
random initializations results in models with poor performance.
 We describe firstly
how we initialize the global parameters of the model $\lambda_{kv}$,
and secondly, the initialization of
topic assignments $\z^{(0)}_d$ when performing
Gibbs sampling on a document.

We test the performance of our model on the CoNLL 2003 NER
dataset~\cite{Hoffart:2011}, a large public dataset for evaluation of
entity annotation systems, and compare to the current best performing
annotation algorithm. 

\subsection{Model initialization and training}\label{sect:data}
An English Wikipedia article is an admissible topic if it
is not a disambiguation, redirect, category or list page, and its main
content section is longer than 50 characters.
Out of the initial 4.1M pages this step selects
3.8M articles, each one defining a topic in the model.
Initial candidate mention strings for a topic are generated from
its title, the 
titles of all Wikipedia pages that redirect to it, and the anchor text of all
its incoming links (within Wikipedia). All mention strings are
lower-cased, single-character mentions are ignored. This amounts
to roughly 11M mention types and 13M mention-topic
parameters. Note, that although this initialization is highly
sparse - for most mention-topic pairs, the initial variational parameter
$\hat{\lambda}_{kv}$ is initialized to zero, this does not
mean that topics cannot be associated with new mentions during training,
due to the inclusion of the prior pseudo-count $\beta$. 

We carry out a minimal filtering
of infrequent and stop words. We compile a list of 600 stop words: all
lower-cased single tokens that occur in more than 5\% of 
Wikipedia. We discard all words that occur in less
than 3 articles. This procedure defines a vocabulary of 1.8M
different (lowercased) words. The total number of topic-word
parameters is approximately 70M.

Let $v$ be a symbol denoting a word or a mention. We
initialize the corresponding parameter $\hat{\lambda}_{kv}$ for
topic $k$ as
$\hat{\lambda}_{kv} = P(k|v) = \frac{\mathrm{count}(v,k)}{\mathrm{count}(v)}$.
For the content words, counts are collected from Wikipedia articles
and for mentions, from titles (including redirects) and anchors. 
For each word we retain in the initial model the top 500 scoring
topics, according to $P(k|v)$. We don't limit the number of
topics associated with mentions. 
Notice that parameters not explicitly represented in the initial
model are still eligible for sampling via the pseudocount
$\beta$, thus the full model
is given by the
cross-product of the vocabularies and the topics.

We train model on the English Gigaword Corpus\footnote{LDC
  Catalogue: 2003T05}, a
collection of news-wire text consisting of
4.8M documents, containing a total of 2 billion tokens.
We annotate the text with an off-the-shelf CoNLL-style named entity
recognizer which identifies mentions of organizations, people
and location names. We ignore the label and simply use
the entity boundaries to identify mention spans.
\begin{figure} \centering
\includegraphics[scale=0.37,trim=0cm 7cm 0cm 7cm]{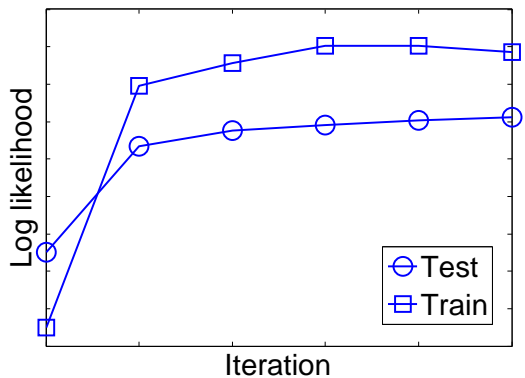}
\caption{Log likelihood on train and held-out data (rescaled to fit on
  same axes).}  
\label{fig:logl}
\end{figure}
Before evaluating in terms of entity disambiguation, as an
objective measure of model quality, we compute the log
likelihood on a held-out set
with a `left-to-right' approximation~\cite{wallach2009}. 
Figure~\ref{fig:logl} shows that the model behaves as
expected and appears not to overfit:
both train and held-out likelihood increase with
each iteration, levelling out over time.

\subsection{Sampler initialization} \label{sect:init}
A naive initialization of the Gibbs sampler
could use the topic with the greatest parameter
value for a word $z_i^{(0)}=\argmax_k\lambda_{kv}$, or even random
assignments.
We find that these are not good solutions.
Poor performance arises because the
distribution of topics for a mention is typically long-tailed. If the
true topic for a mention is not the most likely one, its parameter
value could be several orders of magnitude smaller than the primary topic.
The problem is that topics have extremely fine
granularity and even with sparse priors it is unlikely that
the right patterns of topic mixtures will emerge by brute-force
sampling in a reasonable amount of time.

To improve the initialization we use a simpler, and faster,
heuristic disambiguation algorithm derived from TagMe's
annotator~\cite{Ferragina:2010}. 
The score for topic $z_{k}$ being assigned to mention $w_i$
is defined as the edge-weighted contribution from all other
mentions in the document:
\begin{equation}\label{eq:tagme1}
\mathrm{rel}(z_{k}|w_i) = \sum_{j\neq i} \mathrm{votes}_j(z_{k})\,,\notag
\end{equation}
where the edge-weighted votes are defined as:

\begin{equation}
\mathrm{votes}_j(z_{k})
	=\frac{\sum_{k'\in \bm{z}(w_j)}\mathrm{sim}(z_{k},z_{k'})\lambda_{k'w_j}}{|\bm{z}(w_j)|}\,,
\end{equation}

and ${\bm{z}(v)}$ indexes the set of topics $k$ with
$\hat{\lambda}_{kv}>0$. The similarity measure is that used in
Equation~(\ref{eq:coh}).
Given hyperparameters $\epsilon$ and $\tau$ TagMe
excludes from the set of candidates for mention $w_i$ topics
with score lower than $\max_k \text{rel}(z_{w_ik})\times\epsilon$, and
$P(k|w_i)<\tau$. Within this set the candidate $\argmax_k
\lambda_{kw_i}$ is selected.
Intuitively, this method first selects a set of topics that
are closely related according to the graph, then
picks the one with the highest prior.



\subsection{Evaluation data and metrics}
A well-studied dataset for named entity recognition is the
English CoNLL 2003 NER dataset~\cite{TjongKimSang:2003}, a corpus
of Reuters news annotated with person, location, organization and 
miscellaneous tags. 
It is divided in three
partitions: train (946 documents), test-a (216 documents, used for
development) and test-b (231 documents, used for blind
evaluation). The dataset was augmented 
with identifiers from YAGO, Wikipedia and Freebase to evaluate entity
disambiguation systems~\cite{Hoffart:2011}. We refer to this dataset
as CoNLL-Aida.
In our experiments on this data we report \emph{micro-accuracy}: the fraction
of mentions whose  
predicted topic is the same as the gold-standard annotation.
There are 4,788
mentions in test-a and 4,483  
mentions in test-b. We also report \emph{macro-accuracy}, where
document-level accuracy is averaged by the total number of documents.

\subsection{Hyper-parameters}
Since we don't train on the CoNLL-Aida data, we set
the hyper-parameters of the model by carrying out a greedy search
that optimizes the sum of the micro and macro scores on both the train and
test-a partitions.

Our model has a few hyperparameters,
$\alpha$, $\beta^c,\beta^m$, the number of Gibbs sweeps and the number
of iterations.
We find that comparable performance can be achieved using a wide
range of values.
The priors control the
degree of exploration of the sampler. $\alpha$ acts as a
pseudo-count for each topic in a document. If this parameter is set to
zero the sampler can visit only topics that have already been observed in the
document; although this ensures a high degree of consistency in the topics,
preventing any exploration in this manner is detrimental to performance.
We find that any $\alpha\in [10^{-5},10^{-1}]$ works
well.
$\beta$ provides a residual probability
that any word/mention can be associated with a topic, thus controlling
exploration in sampling and vocabulary growth.
$\beta$ also
regularizes the sampling distribution;
the denominator $\kappa'_{z_i}$ in Equation~\eqref{eq:rearrangement} is
a function of $\beta$. If  
$V\beta$ is too small, the topics with very small parameters
$\lambda$ can be sampled with high probability.
For our model the vocabulary is in the order of $10^6$, thus
 in practice we find $\beta\in[10^{-7},10^{-3})$ works well for both
   words and mentions.

The robustness of the model's performance to these wide range of
hyperparameter settings advocates the use of our probabilistic
approach. Conversely, we find that approaches built upon
heuristic scoring metrics, such as our TagMe-like
algorithm for sampler initialization require much more careful tuning.
We found that $\epsilon$ and $\tau$, values around 0.25
and 0.02, respectively, worked well. 

We obtain the best results after one training iteration, 
this is probably because Wikipedia essentially
provides a (noisy) labeled dataset to
fix the initial parameters, which yields a strong initialization.
Indeed, a number of approaches just use a Wikipedia initialization
like ours alone,
along with the graph. Note, however, that without
running inference in model, the
initial model alone, even with the guidance of the Wikipedia in-link
graph, e.g., as in the TagMe tagging
algorithm, does not yield optimal performance 
(see Table~\ref{tab:results}, column `TagMe'). It is the fast Gibbs
sampler used in combination with the Wikipedia in-link graph
which greatly improves the annotation accuracy.

In terms of Gibbs sweeps the best results
are achieved with 800 sweeps but the improvement over 50
(which we use for training) is marginal.

\subsection{Results}
\begin{table}
\centering
\begin{tabular}{c|p{1cm}|p{1.1cm}|p{1cm}|p{1.8cm}|} \cline{2-5}
&\multicolumn{4}{|c|}{{\bf test-a}} \\\cline{2-5}
           &Base &TagMe*& Aida& Wiki-LDA\\\hline
\multicolumn{1}{|c|}{Micro} &70.76& 76.89& 79.29 &{\bf 80.97}~$\pm$0.49\\\hline
\multicolumn{1}{|c|}{Macro} &69.58& 74.57& 77.00 &{\bf 78.61}\\\cline{1-5}
&\multicolumn{4}{|c|}{{\bf test-b}} \\\cline{1-5}
\multicolumn{1}{|c|}{Micro} &69.82 & 78.64& 82.54 &{\bf 83.71}~$\pm$0.50\\\hline
\multicolumn{1}{|c|}{Macro} &72.74 & 78.21& 81.66 &{\bf 82.88} \\
\hline\end{tabular}
\caption{\label{tab:results}Accuracy on the CoNLL-Aida corpus.}
\end{table}

Table~\ref{tab:results} summarizes the disambiguation evaluation results.
The Baseline predicts for mention $m$ the topic $k$ maximizing
$P(k|m)$. The baseline is quite high, this is due to the
skewed distribution of topics -- which makes the problem
challenging. The second column reports the accuracy of our implementation
of TagMe (TagMe*), used to initialize the sampler. Finally, we compare
against the best of the Aida systems, extensively benchmarked
in~\cite{Hoffart:2011} where they proved superior to all the best
currently published systems. We report figures for the latest
best model (``r-prior sim-k r-coh''), periodically updated by
the Aida
group.\footnote{\url{http://www.mpi-inf.mpg.de/yago-naga/aida/} as of
  May 2013. We 
  thank Johannes Hoffart for providing us with the 
  latest best results on the test-a partition in personal communications.}
The proposed method, Wiki-LDA, has the best results on both development
(test-a) and blind evaluation (test-b). For completeness, we
report micro and macro figures for the train partition:
83.04\% and 82.84\% respectively.
We report standard deviations on the micro accuracy scores of our model
obtained via bootstrap re-sampling of the system's predictions.


Inspection of errors on the development partitions
revealed at least one clear issue.
In some documents, a mention can appear multiple times
with different gold annotations. E.g. in one article,
`Washington' appears multiple times, sometimes annotated as the city,
and sometimes as USA (country); in another,
`Wigan' is annotated both as the UK town, and its rugby club.
Due to the `bag-of-words' assumption,
the Wiki-LDA model is not able to discriminate such cases and
 naturally tends to commit to
one assignment per string per document.
Local context could help disambiguate these cases.
It would be relatively straightforward to up-weight this context in
our sampler; e.g. by weighting the influence of assignments by
a distance function. This extension is left for future work.

\subsection{Efficiency remarks}
The goal of this work is not simply provide a new scalable inference
framework for LDA, but to produce a system sufficiently
scalable to address the entity-disambiguation task effectively,
hence achieving
state-of-the-art performance in this domain.
Indeed, direct comparison to other scalable LDA algorithms is
impossible due to the different regimes in which the models operate --
typical LDA models seek to `compress' the documents, representing
them with a small set of topics, but our model addresses
annotation with a very large number of topics. 
However, we attempt to roughly compare approximate computation
times and memory requirements
with the current state-of-the-art scalable LDA frameworks.

The time needed to train on 5M documents with
50 Gibbs sweeps per document on 1,000 machines is approximately
one hour. The memory footprint is negligible (a few hundred Mb).
As noted, one cannot compare directly to the current
distributed LDA systems 
which use far fewer topics and run different inference algorithms
(usually pure Gibbs sampling), however, some of the fastest systems to date are
reported in \cite{smola2010architecture}. 
This work reports a maximum throughput of around 16k-30k
documents/machine per hour 
on different corpora, using 100 machines, beyond this number they run out of
memory. They use a complex architecture and vanilla LDA, with our
simple architecture and a much (5000 times) larger topic space our
training rates are certainly 
comparable. In addition, in our architecture based on pipelines of
MapReduce speed should, in principle at least, correlate linearly with the
number of machines as the processes run independently of each
other. We plan to investigate these issues further in the 
future.
The LDA model proposed in \cite{han-sun:2012:EMNLP-CoNLL} is somewhat 
comparable, they report a training time of over a week with 20G memory,
on a single machine.

\section{Conclusion and Future Work}
Topic models provide a principled, flexible framework for analyzing
latent structure in text. These are desirable properties for a whole
new area of work that is beginning to systematically explore semantic
grounding with respect to web-scale knowledge bases such as
Wikipedia and
Freebase. We have proposed a conceptually simple, highly
scalable, and reproducible, distributed inference framework 
built upon pipelines of MapReduces for scaling topic models for the
entity disambiguation task, and beyond. We extended the hybrid
SVI/Gibbs sampling framework to a 
distributed setting and incorporated crucial metadata such as the
Wikipedia link graph into the sampler. The model produced, to the
authors' best knowledge, the best results to date on the CoNLL-Aida evaluation
dataset.
Although we address a different task to the usual applications of LDA
(exploratory analysis and structure discovery in text)
and work in a very different parameter domain,
this system is comparable to, or even faster than state of the art learning
systems for vanilla LDA. 
The topic space and parallelization degree are the largest to date.
Further lines of investigation include implementing
more advanced local/global update schedules,
investigating their interaction with sharding
and batching schemes, and the effect on
 computational efficiency and performance.
On the modeling side our first priority is the inclusion of the
segmentation task directly into the model, and exploring hierarchical variants, 
which could provide an alternative way to incorporate
information from the Wikipedia graph. The graphical structure
could even be further refined from data.

\section*{Acknowledgments}
We would like to thank Michelangelo Diligenti, Yasemin Altun, Amr
Ahmed, Alex Smola, Johannes 
Hoffart, Thomas Hofmann, Marc'Aurelio Ranzato and Kuzman Ganchev for
valuable feedback and discussions.

\bibliographystyle{abbrv}
\bibliography{bibliog}

\begin{thebibliography}{10}

\bibitem{bishop2006}
C.~M. Bishop.
\newblock {\em Pattern Recognition and Machine Learning (Information Science
  and Statistics)}.
\newblock Springer-Verlag New York, Inc., 2006.

\bibitem{blei2003}
D.~M. Blei, A.~Y. Ng, and M.~I. Jordan.
\newblock Latent {D}irichlet {A}llocation.
\newblock {\em Journal of Machine Learning Research}, 3:993--1022, 2003.

\bibitem{Chambers:2010}
C.~Chambers, A.~Raniwala, F.~Perry, S.~Adams, R.~R. Henry, R.~Bradshaw, and
  N.~Weizenbaum.
\newblock Flume{J}ava: {E}asy, efficient data-parallel pipelines.
\newblock In {\em Proceedings of the 2010 ACM SIGPLAN conference on Programming
  language design and implementation}, pages 363--375. ACM, 2010.

\bibitem{Chang:2008}
F.~Chang, J.~Dean, S.~Ghemawat, W.~C. Hsieh, D.~A. Wallach, M.~Burrows,
  T.~Chandra, A.~Fikes, and R.~E. Gruber.
\newblock Bigtable: A distributed storage system for structured data.
\newblock {\em ACM Trans. Comput. Syst.}, 26(2):4:1--4:26, 2008.

\bibitem{Cilibrasi:2007}
R.~L. Cilibrasi and P.~M.~B. Vitanyi.
\newblock The {G}oogle {S}imilarity {D}istance.
\newblock {\em IEEE Trans. on Knowl. and Data Eng.}, 19(3):370--383, 2007.

\bibitem{coates2012emergence}
A.~Coates, A.~Karpathy, and A.~Ng.
\newblock Emergence of object-selective features in unsupervised feature
  learning.
\newblock In {\em Advances in Neural Information Processing Systems 25}, pages
  2690--2698, 2012.

\bibitem{Dean:2008}
J.~Dean and S.~Ghemawat.
\newblock Map{R}educe: {S}implified data processing on large clusters.
\newblock {\em Commun. ACM}, 51(1):107--113, 2008.

\bibitem{Erosheva:2004}
E.~A. Erosheva, E.~Fienberg, and J.~Lafferty.
\newblock Mixed-membership models of scientific publications.
\newblock {\em Proceedings of the National Academy of Sciences},
  97(22):11885--11892, 2004.

\bibitem{Ferragina:2010}
P.~Ferragina and U.~Scaiella.
\newblock Tag{M}e: {O}n-the-fly annotation of short text fragments (by
  wikipedia entities).
\newblock In {\em Proceedings of the 19th ACM international conference on
  Information and knowledge management}, pages 1625--1628. ACM, 2010.

\bibitem{Hall:2010:nips}
K.~Hall, S.~Gilpin, and G.~Mann.
\newblock Map{R}educe/{B}igtable for distributed optimization.
\newblock In {\em Advances in Neural Information Processing Systems: Workshop
  on Learning on Cores, Clusters and Clouds}. MIT Press, 2010.

\bibitem{han-sun:2012:EMNLP-CoNLL}
X.~Han and L.~Sun.
\newblock An entity-topic model for entity linking.
\newblock In {\em Proceedings of the 2012 Joint Conference on Empirical Methods
  in Natural Language Processing and Computational Natural Language Learning},
  pages 105--115. Association for Computational Linguistics, 2012.

\bibitem{Hoffart:2011}
J.~Hoffart, M.~A. Yosef, I.~Bordino, H.~F\"{u}rstenau, M.~Pinkal, M.~Spaniol,
  B.~Taneva, S.~Thater, and G.~Weikum.
\newblock Robust disambiguation of named entities in text.
\newblock In {\em Proceedings of the Conference on Empirical Methods in Natural
  Language Processing}, pages 782--792. Association for Computational
  Linguistics, 2011.

\bibitem{hoffman2010}
M.~Hoffman, D.~Blei, and F.~Bach.
\newblock Online learning for {L}atent {D}irichlet {A}llocation.
\newblock {\em Advances in Neural Information Processing Systems}, 23:856--864,
  2010.

\bibitem{hoffman2012}
M.~Hoffman, D.~M. Blei, C.~Wang, and J.~Paisley.
\newblock Stochastic variational inference.
\newblock {\em arXiv preprint arXiv:1206.7051}, 2012.

\bibitem{Kataria:2011}
S.~S. Kataria, K.~S. Kumar, R.~R. Rastogi, P.~Sen, and S.~H. Sengamedu.
\newblock Entity disambiguation with hierarchical topic models.
\newblock In {\em Proceedings of the 17th ACM SIGKDD international conference
  on Knowledge discovery and data mining}, pages 1037--1045. ACM, 2011.

\bibitem{Kulkarni:2009}
S.~Kulkarni, A.~Singh, G.~Ramakrishnan, and S.~Chakrabarti.
\newblock Collective annotation of wikipedia entities in web text.
\newblock In {\em Proceedings of the 15th ACM SIGKDD international conference
  on Knowledge discovery and data mining}, pages 457--466. ACM, 2009.

\bibitem{lafferty2005correlated}
J.~D. Lafferty and D.~M. Blei.
\newblock Correlated topic models.
\newblock In {\em Advances in neural information processing systems}, pages
  147--154, 2005.

\bibitem{Le:2012}
Q.~Le, M.~Ranzato, R.~Monga, M.~Devin, K.~Chen, G.~Corrado, J.~Dean, and A.~Ng.
\newblock Building high-level features using large scale unsupervised learning.
\newblock In {\em Proceedings of the 29th Annual International Conference on
  Machine Learning}. ACM, 2012.

\bibitem{mcdonald-hall-mann:2010:NAACLHLT}
R.~McDonald, K.~Hall, and G.~Mann.
\newblock Distributed training strategies for the structured perceptron.
\newblock In {\em Human Language Technologies: The 2010 Annual Conference of
  the North American Chapter of the Association for Computational Linguistics},
  pages 456--464, Los Angeles, California, June 2010. Association for
  Computational Linguistics.

\bibitem{Mihalcea:2007}
R.~Mihalcea and A.~Csomai.
\newblock Wikify!: {L}inking documents to encyclopedic knowledge.
\newblock In {\em Proceedings of the sixteenth ACM conference on Conference on
  information and knowledge management}, pages 233--242. ACM, 2007.

\bibitem{Milne:2008b}
D.~Milne and I.~Witten.
\newblock An effective, low-cost measure of semantic relatedness obtained from
  {W}ikipedia links.
\newblock In {\em Proceedings of the first AAAI Workshop on Wikipedia and
  Artificial Intelligence ({WIKIAI}'08)}, 2008.

\bibitem{Milne:2008}
D.~Milne and I.~H. Witten.
\newblock Learning to link with {W}ikipedia.
\newblock In {\em Proceedings of the 17th ACM conference on Information and
  knowledge management}, pages 509--518. ACM, 2008.

\bibitem{mimno2012}
D.~Mimno, M.~Hoffman, and D.~Blei.
\newblock Sparse stochastic inference for {L}atent {D}irichlet allocation.
\newblock {\em arXiv preprint arXiv:1206.6425}, 2012.

\bibitem{Newman:2006}
D.~Newman, C.~Chemudugunta, and P.~Smyth.
\newblock Statistical entity-topic models.
\newblock In {\em Proceedings of the 12th ACM SIGKDD international conference
  on Knowledge discovery and data mining}, pages 680--686. ACM, 2006.

\bibitem{Olston:2008}
C.~Olston, B.~Reed, U.~Srivastava, R.~Kumar, and A.~Tomkins.
\newblock Pig latin: {A} not-so-foreign language for data processing.
\newblock In {\em Proceedings of the 2008 ACM SIGMOD international conference
  on Management of data}, pages 1099--1110. ACM, 2008.

\bibitem{ranganath2013adaptive}
R.~Ranganath, C.~Wang, D.~M. Blei, and E.~P. Xing.
\newblock An adaptive learning rate for stochastic variational inference.
\newblock In {\em International Conference on Machine Learning}, 2013.

\bibitem{ratinov2011local}
L.-A. Ratinov, D.~Roth, D.~Downey, and M.~Anderson.
\newblock Local and global algorithms for disambiguation to wikipedia.
\newblock In {\em ACL}, volume~11, pages 1375--1384, 2011.

\bibitem{robbins1951stochastic}
H.~Robbins and S.~Monro.
\newblock A stochastic approximation method.
\newblock {\em The Annals of Mathematical Statistics}, pages 400--407, 1951.

\bibitem{Scaiella:2012}
U.~Scaiella, P.~Ferragina, A.~Marino, and M.~Ciaramita.
\newblock Topical clustering of search results.
\newblock In {\em Proceedings of the fifth ACM international conference on Web
  search and data mining}, pages 223--232. ACM, 2012.

\bibitem{schaul2012no}
T.~Schaul, S.~Zhang, and Y.~LeCun.
\newblock No more pesky learning rates.
\newblock {\em arXiv preprint arXiv:1206.1106}, 2012.

\bibitem{smola2010architecture}
A.~Smola and S.~Narayanamurthy.
\newblock An architecture for parallel topic models.
\newblock {\em Proceedings of the VLDB Endowment}, 3(1-2):703--710, 2010.

\bibitem{teh2007}
Y.~W. Teh, D.~Newman, and M.~Welling.
\newblock A collapsed variational bayesian inference algorithm for latent
  dirichlet allocation.
\newblock {\em Advances in Neural Information Processing Systems}, 19:1353,
  2007.

\bibitem{TjongKimSang:2003}
E.~F. Tjong Kim~Sang and F.~De~Meulder.
\newblock Introduction to the conll-2003 shared task: {L}anguage-independent
  named entity recognition.
\newblock In {\em Proceedings of the seventh conference on Natural language
  learning at HLT-NAACL 2003 - Volume 4}, pages 142--147. Association for
  Computational Linguistics, 2003.

\bibitem{wallach2009}
H.~Wallach, I.~Murray, R.~Salakhutdinov, and D.~Mimno.
\newblock Evaluation methods for topic models.
\newblock In {\em Proceedings of the 26th Annual International Conference on
  Machine Learning}, pages 1105--1112. ACM, 2009.

\bibitem{wang2009}
Y.~Wang, H.~Bai, M.~Stanton, W.-Y. Chen, and E.~Y. Chang.
\newblock Plda: Parallel latent dirichlet allocation for large-scale
  applications.
\newblock In {\em Algorithmic Aspects in Information and Management}, pages
  301--314. Springer, 2009.

\end{thebibliography}

\end{document}